\documentclass[11pt,letterpaper]{article}
\usepackage{emnlp2017}
\usepackage{times}
\usepackage{latexsym}
\usepackage{color}
\usepackage{url}
\usepackage{comment}
\usepackage{subcaption}
\usepackage{diagbox,booktabs}
\usepackage{hyperref}
\hypersetup{draft}
\usepackage{multirow}
\usepackage{amsmath,array,graphicx}
\usepackage{tablefootnote}

\graphicspath{ {figures/} }
\emnlpfinalcopy

\def\emnlppaperid{***}

\title{A Continuously Growing Dataset of Sentential Paraphrases}

\author{Wuwei Lan\textsuperscript{1}, Siyu Qiu\textsuperscript{2}, Hua He\textsuperscript{3}, Wei Xu\textsuperscript{1} \\ \textsuperscript{1} Department of Computer Science and Engineering \\
Ohio State University\\
  {\tt \{lan.105, xu.1265\}@osu.edu}\\
  \textsuperscript{2} Computer and Information Science Department\\
  University of Pennsylvania \\
  {\tt siqiu@seas.upenn.edu} \\
  \textsuperscript{3} Department of Computer Science\\ University of Maryland, College Park\\
  {\tt huah@umd.edu}
}

\date{}

\begin{document}

\maketitle

\begin{abstract}
A major challenge in paraphrase research is the lack of parallel corpora. In this paper, we present a new method to collect large-scale sentential paraphrases from Twitter by linking tweets through shared URLs. The main advantage of our method is its simplicity, as it gets rid of the classifier or human in the loop needed to select data before annotation and subsequent
application of paraphrase identification algorithms in the previous work. We present the largest human-labeled paraphrase corpus to date of 51,524 sentence pairs and the first cross-domain benchmarking for automatic paraphrase identification. In addition, we show that more than 30,000 new sentential paraphrases can be easily and continuously captured every month at $\sim$70\% precision, and demonstrate their utility for downstream NLP tasks through phrasal paraphrase extraction. We make our code and data freely available.\footnote{The code and data can be obtained from the first and last author's websites.}
\end{abstract}

\section{Introduction}

A paraphrase is a restatement of meaning using different expressions \cite{bhagat2013paraphrase}. It is a fundamental semantic relation in human language, as formalized in the Meaning-Text linguistic theory which defines meaning as `invariant of paraphrases' \cite{milicevic2006short}. Researchers have shown benefits of using paraphrases in a wide range of applications \cite{madnani2010generating}, including question answering \cite{fader2013paraphrase}, semantic parsing \cite{berant2014semantic}, information extraction \cite{sekine2006demand,zhang2015exploiting}, machine translation \cite{mehdizadehseraj-siahbani-sarkar:2015:EMNLP}, textual entailment \cite{dagan2006pascal,bjerva2014meaning,marelli2014semeval,izadinia2015segment}, vector semantics \cite{faruqui2015retrofitting,wieting-15}, and semantic textual similarity \citep{agirrea2015semeval,li-srikumar:2016:EMNLP2016}. Studying paraphrases in Twitter can also help track unfolding events \cite{vosoughi2016semi} or the spread of information \cite{bakshy2011everyone} on social networks.  

\begin{table*}[!ht]
\small
\centering
\begin{tabular}{lccccc}
\hline 
\multicolumn{1}{c}{\bf Name} &\bf Genre &\bf Size &\bf  Sentence Length &\bf Multi-Ref. &\bf Non-Para. \\ 
\hline
MSR Paraphrase Corpus (MSRP) & news & 5801 pairs & 18.9 words & no & yes \\
Twitter Paraphrase Corpus (PIT-2015) & Twitter & 18,762 pairs & 11.9 words & some & yes \\
Twitter News URL Corpus (this work) & Twitter &  44,365 pairs & 14.8 words & yes & yes \\
\hline
 MSR Video Description Corpus & YouTube & 123,626 sentences & 7.03 words & yes & no \\
\hline
\end{tabular}
\caption{\label{tab:survey} Summary of publicly available large sentential paraphrase corpora with manual quality assurance. Our Twitter News URL Corpus has the advantages of including both meaningful non-paraphrases (Non-Para.) and multiple references (Multi-Ref.), which are important for training paraphrase identification and evaluating paraphrase generation, respectively.}
\vspace{-.2in}
\end{table*}

In this paper, we address a major challenge in paraphrase research --- the lack of parallel corpora. There are \textbf{only two} publicly available datasets of naturally occurring sentential paraphrases and non-paraphrases:\footnote{Meaningful non-paraphrases (pairs of sentences that have similar wordings or topics but different meanings, and that are not randomly or artificially generated) have been very difficult to obtain but are very important, because they serve as necessary distractors in training and evaluation.} the MSRP corpus derived from clustered news articles \cite{dolan2005automatically} and the PIT-2015 corpus from Twitter trending topics \cite{Xu-EtAl-2014:TACL,xu2015semeval}. Our goal is not only to create a new annotated paraphrase corpus, but to identify a new data source and method that can narrow down the search space of paraphrases without using the classifier-biased or human-in-the-loop data selection as in MSRP and PIT-2015. This is so that sentential paraphrases can be conveniently and continuously harvested in large quantities to benefit downstream applications. 

We present an effective method to collect sentential paraphrases from tweets that refer to the same URL and contribute a new gold-standard annotated corpus of 51,524 sentence pairs, which is the largest to date (Table \ref{tab:survey}). We show the different characteristics of this new dataset contrasting the two existing corpora through the first systematic study of paraphrase identification across multiple datasets. Our new corpus is complementary to previous work, as the corpus contains multiple references of both formal well-edited and informal user-generated texts. This is also the first work that provides a continuously growing collection, with more than 30,000 new sentential paraphrases per month automatically labeled at $\sim$70\% precision. We demonstrate that up-to-date phrasal paraphrases can then be extracted via word alignment (see examples in Table \ref{tab:phrase_sample}). We plan to continue collecting paraphrases using our method and release a constantly updating paraphrase resource. 





\begin{table}[!ht]
\small
\centering
{%
\begin{tabular}{|m{7cm}|}
\hline
a 15-year-old girl, a 15yr old, a 15 y/o girl \\
\hline
fetuses, fetal tissue, miscarried fetuses \\
 \hline
responsible for, guilty of, held liable for, liable for \\
\hline
UVA administrator, UVa official, U-Va. dean, University of Virginia dean \\
\hline
FBI Director backs CIA finding, FBI agrees with CIA, FBI backs CIA view, FBI finally backs CIA view, FBI now backs CIA view, FBI supports CIA assertion, FBIClapper back CIA's view, The FBI backs the CIA's assessment, FBI Backs CIA, \\
\hline
Donald Trump, DJT, Mr Trump, Donald @realTrump, D*nald Tr*mp, Comrade \#Trump, GOPTrump, Pres-elect Trump, President-Elect Trump, President-elect Donald J. Trump, PEOTUS Trump,  He-Who-Must-Not-Be-Named\tablefootnote{Another 12 name variations are omitted in the paper due to their offensive nature.}\\

\hline

\end{tabular}
}
\caption{\label{tab:phrase_sample} Up-to-date phrasal paraphrases automatically extracted from Twitter with our new method.}
\vspace{-.2in}
\end{table}

\section{Existing Paraphrase Corpora and Their Limitations}
\label{sec:existingcorpora}




To date, there exist only two publicly available corpora of both sentential paraphrases and non-paraphrases:

\paragraph{MSR Paraphrase Corpus [MSRP]} \cite{dolan04,dolan2005automatically} This corpus contains 5,801 pairs of sentences from news articles, with 4,076 for
training and the remaining 1,725 for testing. It was created from clustered news articles by using an SVM classifier (using features including string similarity and WordNet synonyms) to gather likely paraphrases, then annotated by human on semantic equivalence. 
The MSRP corpus has a known deficiency skewed toward over-identification \cite{das2009paraphrase}, because the ``purpose was not to evaluate the potential effectiveness of the classifier itself, but to
identify a reasonably large set of both positive
and plausible `near-miss' negative examples'' \cite{dolan2005automatically}. It contains a large portion of sentence pairs with many ngrams shared in common.

\paragraph{Twitter Paraphrase Corpus [PIT-2015]} \cite{Xu-EtAl-2014:TACL,xu2015semeval} This corpus was derived from Twitter's trending topic data. The training set contains 13,063 sentence pairs on 400 distinct topics, and the test set contains 972 sentence pairs on 20 topics. As numerous Twitter users spontaneously talk about varied topics, this dataset contains many lexically divergent paraphrases. However, this method requires a manual step of selecting topics to ensure the quality of collected paraphrases, because many topics detected automatically are either incorrect or too broad. For example, the topic ``New York'' relates to tweets with a wide range of information and cannot narrow the search space down enough for human annotation and the subsequent
application of classification algorithms.

\begin{table*}[!htbp]
\small
\centering
\begin{tabular}{|p{0.9in}p{5.1in}|}
\hline
\multicolumn{2}{|l|}{\textbf{Twitter News URL Corpus}} \\
\hline
\centering \textit{Original Tweet}  & Samsung halts production of its Galaxy Note 7 as battery problems linger \\
\hline
  & \#Samsung temporarily suspended production of its Galaxy \#Note7 devices following reports  \\

\cline{2-2}
   &
News hit that @Samsung is temporarily halting production of the \#GalaxyNote7. \\
\cline{2-2} \centering \textit{Paraphrase}
   & Samsung still having problems with their Note 7 battery overheating. Completely halt production. \\ 
\cline{2-2}
   &  SAMSUNG HALTS PRODUCTS OF GALAXY NOTE 7 . THE BATTERIES ARE * STILL * EXPLODING . \\

\hline
  & in which a phone bonfire in 1995--a real one--is a metaphor for samsung's current note 7 problems   \\ 
   
\cline{2-2}
\centering \textit{Non-Paraphrase}   & samsung decides, ``if we don't build it, it won't explode.''  
 \\
\cline{2-2}
 &  Samsung's Galaxy Note 7 Phones AND replacement phones have been going up in flames due to the defective batteries \\



\hline
\end{tabular}
\caption{\label{tab:content_sample} A representative set of tweets linked by a URL originated from news agencies (this work). 
}
\end{table*}

\begin{table*}[!htbp]
\centering
\small
\begin{tabular}{|p{1.2in}p{4.7in}|}
\hline
& 1 dasviness louistomlinson overhears harrystyles on the phone\\
\cline{2-2}
\centering \textbf{Twitter Streaming} & when she likes tall guys ??? ??? vine by justjamiie\\
\cline{2-2}
\centering \textbf{URL Data} & shineeasvines jonghyun when he wears shoe lifts \\
\cline{2-2}
& idaliaorellana kimmvanny ladyfea 21 hahaha if he does it he needs heels\\

\hline
\end{tabular}
\caption{\label{tab:streaming_sample} A representative set of tweets linked by a URL in streaming data (generally poor readability).}
\end{table*}

\section{Constructing the Twitter URL Paraphrase Corpus}
For paraphrase acquisition, it has been crucial to find a simple and effective way to locate paraphrase candidates (see related work in Section \ref{sec: related work}). We show the efficacy of tracking URLs in Twitter. This method does not rely on automatic news clustering as in MSRP or topic detection as in PIT-2015, but it keeps collecting good candidate paraphrase pairs in large quantities.

\subsection{Data Source: News Tweets vs. Streaming}

We extracted the embedded URL in each tweet and used Twitter's Search API to retrieve all tweets that contain the same URL. Some tweets use shortened URLs, which we resolve as full URLs. We tracked 22 English news accounts in Twitter to create the paraphrase corpus in this paper (see examples in Table \ref{tab:content_sample}). We will extend the corpus to include other languages and domains in future work. 

As shown in Table \ref{tab:url_num}, nearly all the tweets posted by news agencies have embedded URLs. About 51.17\% of posts contain two URLs, usually one pointing to a news article and the other to media such as a photo or video. Although close to half of the tweets in Twitter streaming data\footnote{We used Twitter’s Streaming API which provided a real-time stream of public tweets posted on Twitter.
} contain at least one URL, most of them are very hard to read (see examples in Table \ref{tab:streaming_sample}).

\begin{table}[!ht]
\centering
\small
\begin{tabular}{c|ccc}
\hline
\bf Data Source & \bf tweets & \bf avg \#url & \bf avg \# url\\
& \bf w/o url & \bf per tweet  & \bf (news) \\
\cline{1-3}
Streaming Data & 55.8\% & 0.52 & \bf per tweet \\ 
\hline
@nytimes & 	1.2\% & 	1.31 & 	0.988 \\
@cnnbrk & 0.0\%	& 1.17 	 & 1\\
@BBCBreaking &1.0\% & 	1.32 	& 	0.99  \\
@CNN & 0.0\%	&1.85 	&1 \\
@ABC & 1.7\% &	1.26 &0.983\\
@NBCNews &	1.1\%	&1.63  &0.989\\
\hline
\end{tabular}
\caption{\label{tab:url_num} Statistics of tweets in Twitter's streaming data and news account data. Many tweets contain more than one URL because media such as photo or video is also represented by URLs. }
\end{table}

\subsection{Filtering of Retweets}

Retweeting is an important feature in Twitter. There are two types: automatic and manual retweets. An automatic retweet is done by clicking the retweet button on Twitter and is easy to remove using the Twitter API. A manual retweet occurs when the user creates a new tweet by copying and pasting the original tweet and possibly adding some extras, such as hashtags, usernames or comments. It is crucial to remove these redundant tweets with minor variations, which otherwise represent a significant portion of the data (Table \ref{tab:tws_num}). We preprocessed the tweets using a tokenizer\footnote{\url{http://www.cs.cmu.edu/~ark/TweetNLP/}} \cite{gimpel2011part} and an in-house sentence splitter. We then filtered out manual retweets using a set of rules, checking if one tweet was a sub- string of the other, or if it only differed in punctuation, or the contents of the ``twitter:title'' or ``twitter:description'' tag in the linked HTML file of the news article. 

Table \ref{tab:tws_num} shows the effectiveness of the filtering. We used PINC, a standard paraphrase metric, to measure ngram-based dissimilarity \cite{chen11}, and Jaccard metric to measure token-based string similarity \cite{jaccard1912distribution}. After filtering, the dataset contains tweets with more significant rephrasing as indicated by higher PINC and lower Jaccard scores.

\begin{table}[!htbp]
\centering
\resizebox{\columnwidth}{!}{%
\begin{tabular}{c|ccc}
\hline
& \bf{avg \#tweets (STD)}  & \bf{avg PINC}& \bf{avg Jaccard} \\ 
\hline
\bf{before filtering} &205.51 (219.66) &0.6153&0.3635 \\ 
 \bf{after filtering} &74.75 (94.39) &0.7603&0.2515\\ 
\hline
\end{tabular}
}
\caption{\label{tab:tws_num} Impact of filtering of manual retweets. }
\end{table}

\subsection{Gold Standard Corpus}

To get the gold-standard paraphrase corpus, we obtained human labels on Amazon Mechanical Turk. We showed annotators an original sentence, and asked them to select sentences with the same meaning from 10 candidate sentences. For each question, we recruited 6 annotators and paid \$0.03 to each worker.\footnote{The low pricing helps to not attract spammers to this easy-to-finish task. We gave bonus to workers based on quality and the average hourly pay for each worker is about \$7.} On average, each question took about 53 seconds to finish. For each sentence pair, we aggregated the paraphrase and non-paraphrase labels using the majority vote. 

We constructed the largest gold standard paraphrase corpus to date, with 42,200 tweets of 4,272 distinct URLs annotated in the training set and 9,324 tweets of 915 distinct URLs in the test set. The training data was collected between 10/10/2016 and 11/22/2016, and testing data between 01/09/2017 and 01/19/2017. In Section \ref{sec:experiment_gold}, we contrast the characteristics of our data against existing paraphrase corpora. 

\paragraph{Quality Control} We evaluated the annotation quality of each worker using Cohen's kappa agreement \cite{artstein2008inter} against the majority vote of other workers. We asked the best workers (the top 528 out of 876) to label more data by republishing the questions done by workers with low reliability (Cohen's kappa \textless 0.4). 

\paragraph{Inter-Annotator Agreement}
In addition, we had 300 sampled sentence pairs independently annotated by an expert. The annotated agreement is 0.739 by Cohen's kappa between the expert and the majority vote of 6 crowdsourcing workers. If we assume the expert annotation is gold, the precision of worker vote is 0.871, the recall is 0.787, and F1 is 0.827, similar to those of PIT-2015.

\subsection{Continuous Harvesting of Sentential Paraphrases}

Since our method directly applies to raw tweets, it can 
continuously extract sentential paraphrases from Twitter. 
In Section \ref{sec:experiment_gold}, we show that this approach can produce a silver-standard paraphrase corpus at about 70\% precision that grows by more than 30,000 new sentential paraphrases {\em per month}.  Section \ref{sec:experiment_silver} presents experiments demonstrating the utility of these automatically identified sentential paraphrases.

\section{Comparison of Paraphrase Corpora}
\label{sec:experiment_gold}

Though paraphrasing has been widely studied, supporting analyses and experiments have thus far often only been conducted on a single dataset. In this section, we present a comparative analysis of our newly constructed gold-standard corpus with two existing corpora by 1) individually examining the instances of
paraphrase phenomena and 2) benchmarking a range of automatic paraphrase identification approaches. 


\begin{table}[!bp]
\centering
\small
\begin{tabular}{c|ccc}
\hline
 per sentence & \bf{MSRP}& \bf{PIT-2015} &\bf{URL}  \\ 
\hline
\bf{Elaboration} & 0.60 & 0.23  & \textbf{0.79}\\  
\bf{Spelling} & 0.17 & 0.13 & \textbf{0.35}\\
\bf{Synonym} & \textbf{0.26} & 0.10 & 0.13\\
\bf{Phrasal} & 0.42 & \textbf{0.56} & 0.35\\
\bf{Anaphora} & 0.27 & 0.08 & \textbf{0.33}\\
\bf{Reordering} & \textbf{0.53} & 0.33 & 0.49\\
\hline
\hline adjusted by sent length*
  & \bf{MSRP*}& \bf{PIT-2015*} &\bf{URL}  \\ 
\hline
\bf{Elaboration} & 0.42 & 0.36  & \textbf{0.79}\\  
\bf{Spelling} & 0.12 & 0.21 & \textbf{0.35}\\
\bf{Synonym} & \textbf{0.18} & 0.16 & 0.13\\
\bf{Phrasal} & 0.29 & \textbf{0.89} & 0.35\\
\bf{Anaphora} & 0.19 & 0.13 & \textbf{0.33}\\
\bf{Reordering} & 0.37 & \textbf{0.52} & 0.49\\
\hline
\end{tabular}

\caption{Mean number of instances of paraphrase phenomena per sentence pair across three corpora.}
\label{tab:paraphrase_phenomena_compare}
\end{table}

\subsection{Paraphrase Phenomena}
In order to show the differences across these three datasets, we sampled 100 sentential paraphrases from each training set and counted occurrences of each
phenomenon in the following categories: Elaboration (textual pairs can differ in total information content, such as \textit{Trump's ex-wife Ivana} and \textit{ Ivana Trump}), Phrasal (alternates of phrases, such as \textit{ taking over } and \textit{replaces}), Spelling (spelling variants, such as \textit{Trump} and \textit{Trumpf}), Synonym (such as \textit{said} and \textit{told}), Anaphora (a full noun phrase in one sentence that corresponds to the counterpart, such as \textit{@MarkKirk} and \textit{Kirk}) and Reordering (when a word, phrase or the whole sentence reorders, or even logically reordered, such as \textit{Matthew Fishbein questioned him} and \textit{under questioning by Matthew Fishbein}). We report the average number of occurrences of
each paraphrase type per sentence pair for each corpus in Table \ref{tab:paraphrase_phenomena_compare}. As sentences tend to be longer in MSRP and shorter in PIT-2015, we also normalized the numbers by the length of sentences to be more comparable to the URL dataset.

These three datasets exhibit distinct and complementary compositions of paraphrase phenomena. MSRP has more synonyms, because authors of different news articles may use different and rather sophisticated words. PIT-2015 contains many phrasal paraphrases, probably due to the fact that most tweets under the same trending topic are written spontaneously and independently. Our URL dataset shows more elaboration, spelling and anaphora paraphrase phenomena, showing that many URL-embedded tweets are created by users with a conscious intention to rephrase the original news headline.


\subsection{Automatic Paraphrase Identification}
\label{sec:benchmark}

We provide a benchmark on paraphrase identification to better understand various models, as well as the characteristics of our new corpus compared to the existing ones. We focus on binary classification of paraphrase/non-paraphrase, and report the maximum F1 measure of any point on the precision-recall curve. 

\subsubsection{Models}
We chose several  representative technical approaches for automatic paraphrase identification:

\paragraph{GloVe}
~\cite{pennington2014glove} This is a word representation model trained on aggregated global word-word co-occurrence statistics from a corpus. We used 300-dimensional word vectors trained on Common Crawl and Twitter, summed the vectors for each sentence, and computed the cosine similarity.

\paragraph{LR}  The logistic regression (LR) model incorporates 18 features based on 1-3 gram overlaps between two sentences ($s_1$ and $s_2$)~\cite{das2009paraphrase}. The features are of the form precision$_n$ (number of n-gram matches divided by the number of n-grams in $s_1$), recall$_n$ (number of n-gram matches divided by the number of n-grams in $s_2$), and F$_n$ (harmonic mean of recall and precision). The model also includes lemmatized versions of these features. 

\begin{table}[ht!]
\vspace{.2in}
\small
\centering
\resizebox{\columnwidth}{!}{%
\begin{tabular}{l|ccc}
\hline
\bf Method &  \bf F1 &\bf Precision&\bf Recall\\ 
\hline

Random  & 0.799 & 0.665 & 1.0\\
Edit Distance  & 0.799 & 0.666 & 1.0\\ 
\hline
GloVe   & 0.812 & 0.707 & 0.952\\
LR   & 0.829 & 0.741 & 0.941\\
WMF (vec) & 0.817 & 0.713 & 0.956\\
LEX-WMF (vec)& \textbf{0.836} & \textbf{0.751} & \textbf{0.943}\\
OrMF (vec)	&  0.820 & 0.733 & 0.930\\
LEX-OrMF (vec) & \textbf{0.833} & \textbf{0.741} & \textbf{0.950}\\
WMF (sim) &  0.812 & 0.728 & 0.918\\
LEX-WMF (sim)& 0.831 & 0.732 & 0.962\\
OrMF (sim)  & 0.815 & 0.699 & 0.976\\
LEX-OrMF (sim)  & 0.832 & 0.735 & 0.958\\
MultiP &  0.800 & 0.667 & 0.998\\
DeepPairwiseWord  & \textbf{0.834} & \textbf{0.763} & \textbf{0.919}\\
\hline
\end{tabular}
}
\caption{ \label{tab:msrp_result} Paraphrase models in the MSR Paraphrase Corpus (MSRP). The bold font in the table represents top three models in the dataset.}
\end{table}

\begin{table}[ht!]
\vspace{.1in}
\small
\centering
\resizebox{\columnwidth}{!}{%
\begin{tabular}{l|ccc}
\hline
\bf Method &\bf F1 &\bf Precision&\bf Recall\\
\hline
Random & 0.346 & 0.209 & 1.0 \\
Edit Distance &  0.363 & 0.236 & 0.789\\ 
\hline
GloVe  &  0.484 & 0.396 & 0.617\\
LR  &  0.645 & 0.669 & 0.623\\
WMF (vec) &  0.594 & 0.681 & 0.526\\
LEX-WMF (vec)&  0.635 & 0.655 & 0.617\\
OrMF (vec)	&  0.594 & 0.681 & 0.526\\
LEX-OrMF (vec)  & 0.638 & 0.579 & 0.709\\
WMF (sim) & 0.553 & 0.570 & 0.537\\
LEX-WMF (sim) &  \textbf{0.651} & \textbf{0.657} & \textbf{0.646}\\
OrMF (sim)  & 0.563 & 0.591 & 0.537\\
LEX-OrMF (sim)  & 0.644 & 0.632 & 0.657\\
MultiP  & \textbf{0.721} & \textbf{0.705} & \textbf{0.737}\\
DeepPairwiseWord  & \textbf{0.667} & \textbf{0.725} & \textbf{0.617}\\
\hline
\end{tabular}
}
\caption{ \label{tab:twitter_trending_topic_result} Paraphrase models in the Twitter Paraphrase Corpus (PIT-2015).}
\end{table}

\begin{table}[ht!]
\vspace{.1in}
\small
\centering
\resizebox{\columnwidth}{!}{%
\begin{tabular}{l|ccc}
\hline
\bf Method &  \bf F1 &\bf Precision&\bf Recall\\ 
\hline
Random &  0.327 & 0.195 & 1.000\\
Edit Distance  & 0.526 & 0.650 & 0.442\\ 
\hline
GloVe   & 0.583 & 0.607 & 0.560\\
LR  &  0.683 & 0.669 & 0.698\\
WMF (vec) &  0.660 & 0.640 & 0.680\\
LEX-WMF (vec) & \textbf{0.693} & \textbf{0.687} & \textbf{0.698}\\
OrMF (vec)	&  0.662 & 0.625 & 0.703\\
LEX-OrMF (vec) &  \textbf{0.691} & \textbf{0.709} & \textbf{0.674}\\
WMF (sim) &  0.659 & 0.595 & 0.738\\
LEX-WMF (sim) & 0.688 & 0.632 & 0.754\\
OrMF (sim) &  0.660 & 0.690 & 0.632\\
LEX-OrMF (sim)  & 0.688 & 0.630 & 0.758\\
MultiP & 0.536 & 0.386 & 0.875\\
DeepPairwiseWord  & \textbf{0.749} & \textbf{0.803} & \textbf{0.702}\\
\hline
\end{tabular}
}
\caption{ \label{tab:twitter_news_url_result} Paraphrase models in Twitter URL Corpus (this work). }
\end{table}

\begin{figure*}[!ht]
\centering
\begin{subfigure}{.323\linewidth}
\includegraphics[width=\linewidth]{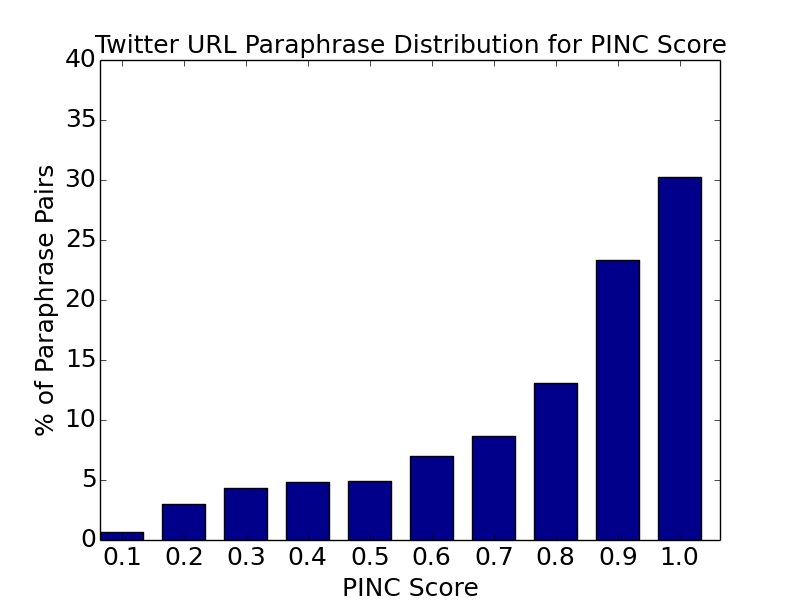}
\caption{Twitter URL}\label{fig:pinc_url}
\end{subfigure}
\begin{subfigure}{.323\linewidth}
\includegraphics[width=\linewidth]{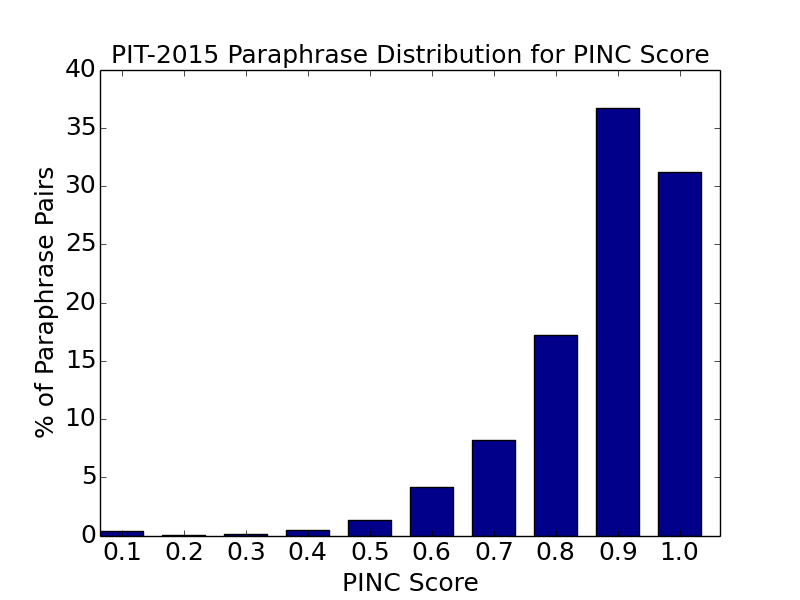}
\caption{PIT-2015}\label{fig:pinc_pit}
\end{subfigure}
\begin{subfigure}{.323\linewidth}
\includegraphics[width=\linewidth]{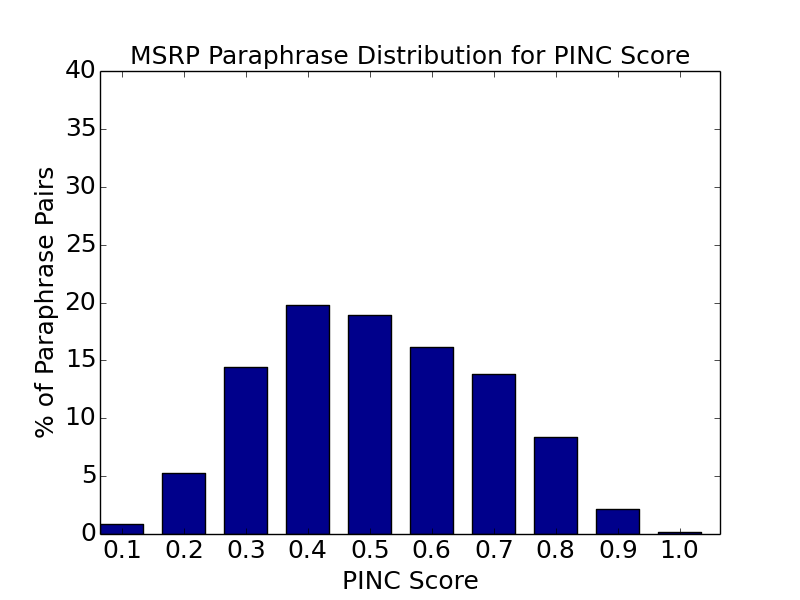}
\caption{MSRP}\label{fig:pinc_msrp}
\end{subfigure}
\caption{Comparison of ngram dissimilarity (PINC score) in sentential paraphrases across three corpora. The MSRP contains sentential paraphrases with more ngram overlaps (low PINC). Our URL corpus and PIT-2015 contain more lexically divergent paraphrases (high PINC).}


\label{fig:pinc_compare}
\end{figure*}

\paragraph{WMF/OrMF} 
Weighted Matrix Factorization (WMF) \cite{guo2012modeling} is an unsupervised latent space model. The unobserved words are carefully handled, which results in more robust embeddings for short texts. Orthogonal Matrix Factorization (OrMF) \cite{guo2014fast} is the extension of WMF, with an additional objective to obtain nearly orthogonal dimensions in matrix factorization to discount redundant information. Specifically, for the \textbf{(vec)} version, vectors of a pair of sentences $\vec{v_1}$ and $\vec{v_2}$ are converted into one feature vector, $[\vec{v_1}+\vec{v_2}, |\vec{v_1}-\vec{v_2}|]$, by concatenating the element-wise sum $\vec{v_1}+\vec{v_2}$ and absolute difference $|\vec{v_1}-\vec{v_2}|$. We also provide the \textbf{(sim)} variation, which directly uses the single cosine similarity score between two sentence vectors.

\paragraph{LEX-WMF/LEX-OrMF} This is an open-sourced adaptation \cite{Xu-EtAl-2014:TACL} of LEXDISCRIM \cite{jidiscriminative} that have shown comparable performance. It combines WMF/OrMF with n-gram overlapping features to train a LR classifier. 

\paragraph{MultiP}
MultiP \cite{Xu-EtAl-2014:TACL} is a multi-instance learning model suited for short messages on Twitter. The at-least-one-anchor assumption in this model looks for two sentences that have a topical phrase in common, plus at least one pair of anchor words that carry a similar key meaning. This model achieved the best performance in the PIT-2015 \cite{Xu-EtAl-2014:TACL} dataset.

\paragraph{DeepPairwiseWord}
He et al. \shortcite{he-lin:2016:N16-1} developed a deep neural network model that focuses on important pairwise word interactions across input sentences. This model innovates in proposing a similarity focus layer and a 19-layer very deep convolutional neural network to guide model attention to important word pairs. It has shown state-of-the-art performance on several textual similarity measurement datasets.

\subsubsection{Model Performance and Dataset Difference}
\label{sec:model_analysis}

The results on three benchmark paraphrase corpora are shown in Table \ref{tab:msrp_result}, \ref{tab:twitter_trending_topic_result} and \ref{tab:twitter_news_url_result}. The random baseline reflects that close to 80\% sentence pairs are paraphrases in the MSPR corpus. This is atypical in the real-world text data and may cause falsely positive predictions. 

Both the edit distance and the LR models exploit surface word features. In particular, the LR model that uses lemmatization and ngram overlap features achieves very competitive performance on all datasets. Figure \ref{fig:pinc_compare} shows a closer look at ngram differences across datasets measured by the PINC metric \cite{chen11}, which is the opposite of BLEU \cite{papineni-EtAl:2002:ACL}. MSRP consists of paraphrases with more ngram overlap (lower PINC), while PIT-2015 contains shorter and more lexically  dissimilar sentences. Our new URL corpus is in between the two, and is more similar to PIT-2015. It includes user's intentional rephrasing of an original tweet from a news agency with some words untouched, as well as some dramatic paraphrases that are challenging for any automatic identification methods, such as \textit{CO2 levels mark `new era' in the world's changing climate} and \textit{CO2 levels haven't been this high for 3 to 5 million years}.

\begin{figure*}[!ht]
\centering
\begin{subfigure}{.323\linewidth}
\includegraphics[width=\linewidth]{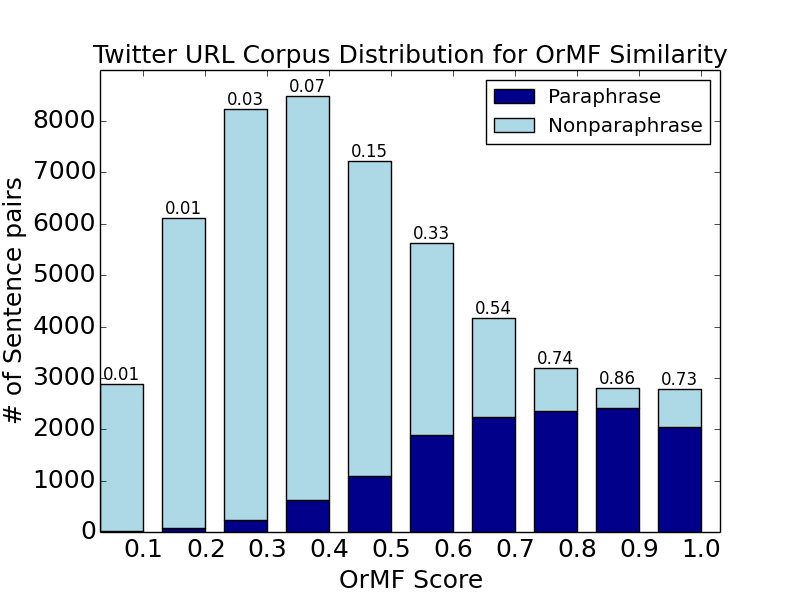}
\caption{Twitter URL}\label{fig:OrMF_url}
\end{subfigure}
\begin{subfigure}{.323\linewidth}
\includegraphics[width=\linewidth]{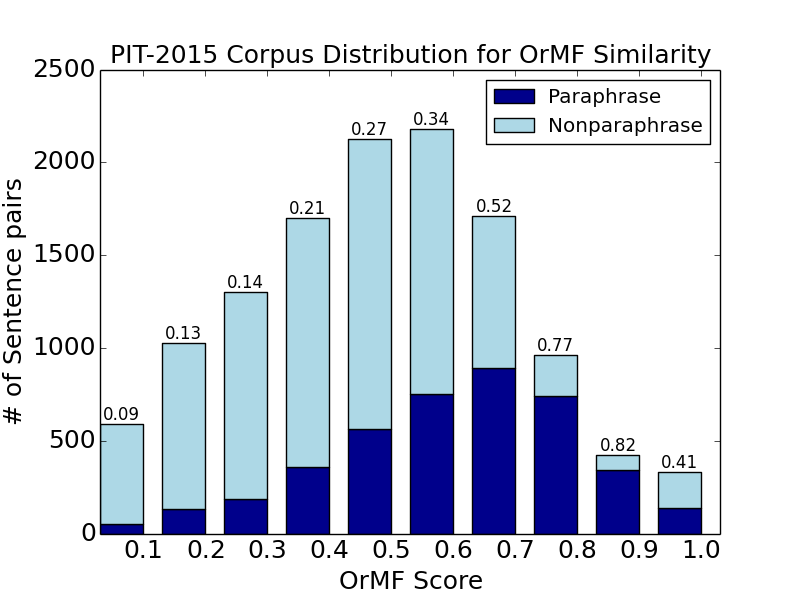}
\caption{PIT-2015}\label{fig:OrMF_pit}
\end{subfigure}
\begin{subfigure}{.323\linewidth}
\includegraphics[width=\linewidth]{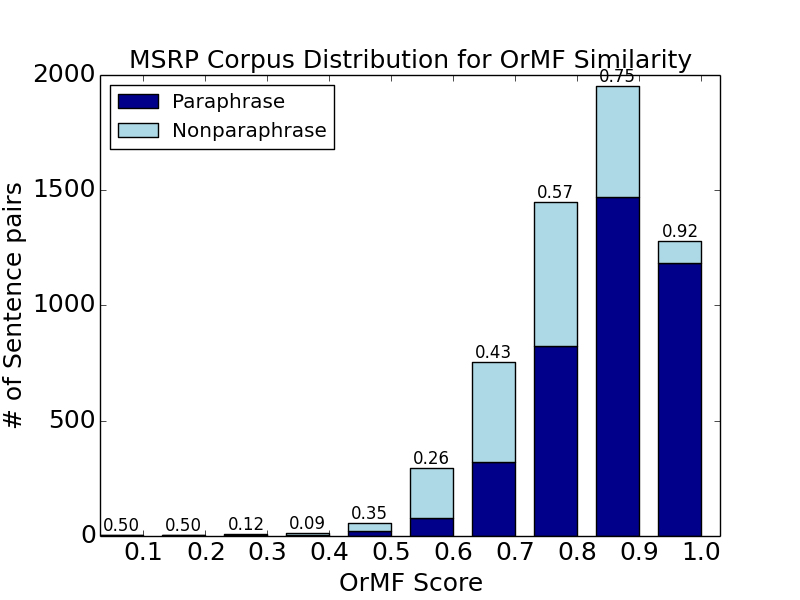}
\caption{MSRP}\label{fig:OrMF_msrp}
\end{subfigure}
\caption{Comparison of OrMF-based distributional semantic similarity across three paraphrase corpora.} 

\label{fig:OrMF_compare}
\end{figure*}

MultiP exploits a restrictive constraint that the candidate sentence pairs share a same topical phrase. It achieves the best performance on PIT-2015, which naturally contains such phrases. For MSRP and URL datasets, we uses the named entity tagged with the longest span as an approximation of a shared topic phrase and thus suffered a performance drop.

Both Glove and WMT/OrMF utilize the underlying co-occurrence statistics of the text corpus. WMT/OrMF use global matrix factorization to project sentences into lower dimension and show great advantages on measuring sentence-level semantic similarities over Glove, which focuses on word representations. Figure \ref{fig:OrMF_compare} shows that the fine-grained distribution of the OrMF-based cosine similarities and that the URL-linked Twitter data works well with OrMF to yield sentential paraphrases. Once combined with ngram overlap features, LEX-WMF and LEX-OrMF show consistently high performance across different datasets, close to the more complicated DeepPairwiseWord. The similarity
focus mechanism on important pairwise word interactions in DeepPairwiseWord is more helpful for the two Twitter datasets, due to the fact that they contain lexically divergent paraphrases while MSRP has an artificial bias toward sentences with high n-gram overlap.

\section{Extracting Phrasal Paraphrases}
\label{sec:experiment_silver}

We can apply paraphrase identification models trained on our gold standard corpus to unlabeled Twitter data and continuously harvest sentential paraphrases in large quantities. We used the open-sourced LEX-OrMF model and obtained 114,025 sentential paraphrases (system predicted probability $\geq$ 0.5 and average precision $=$ 69.08\%) from raw 1\% free Twitter data between 10/10/2016 and 01/10/2017. To demonstrate the utility, we show that we can extract up-to-date lexical and phrasal paraphrases from this data. 

\subsection{Phrase Extraction and Ranking}
 One of the most successful ideas to obtain lexical and phrasal paraphrases in large quantities is through word alignment, then ranking for better quality. This approach was proposed by Bannard \cite{bannard2005paraphrasing} and previously applied to bilingual parallel data to create PPDB \cite{ganitkevitch2013ppdb,PavlickEtAl-2015:ACL:Semantics}. There has been little previous work utilizing monolingual parallel data to learn paraphrases since it is not as naturally available as bitexts.


\begin{figure*}[!ht]
\centering
\begin{subfigure}{.24\linewidth}
\includegraphics[width=\linewidth]{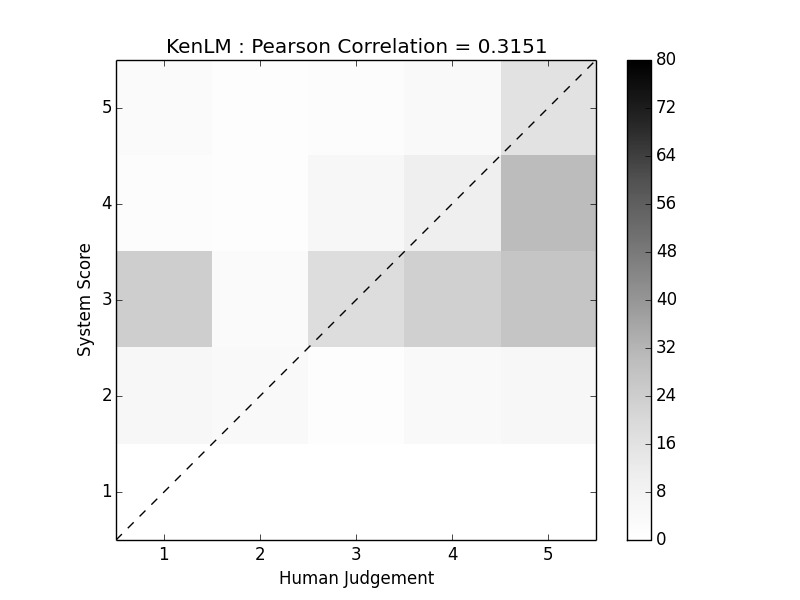}
\caption{Language Model Score \\ \centering{($\rho$ = 0.3151)}}
\label{fig:pearson_lm_score} 
\end{subfigure}
\begin{subfigure}{.24\linewidth}
\includegraphics[width=\linewidth]{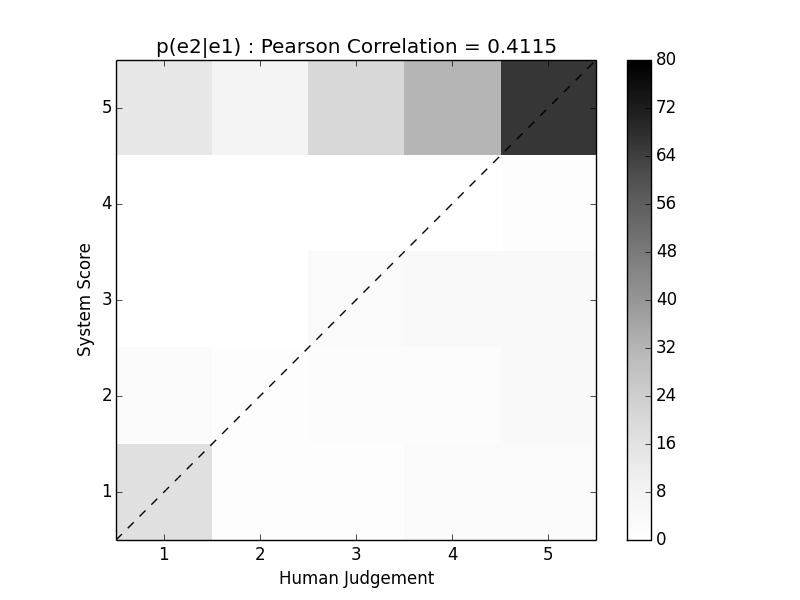}
\caption{Translation Score \\ \centering{($\rho$ = 0.4115)}}
\label{fig:pearson_translation_score} 
\end{subfigure}
\begin{subfigure}{.24\linewidth}
\includegraphics[width=\linewidth]{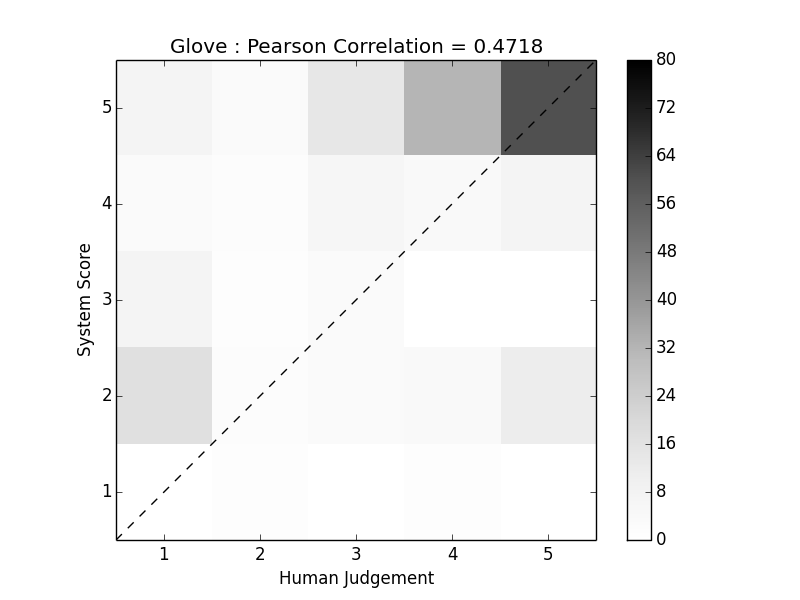}
\caption{Glove Score \\ 
 ($\rho$ = 0.4718)}
\label{fig:pearson_glove_score} 
\end{subfigure}
\begin{subfigure}{.24\linewidth}
\includegraphics[width=\linewidth]{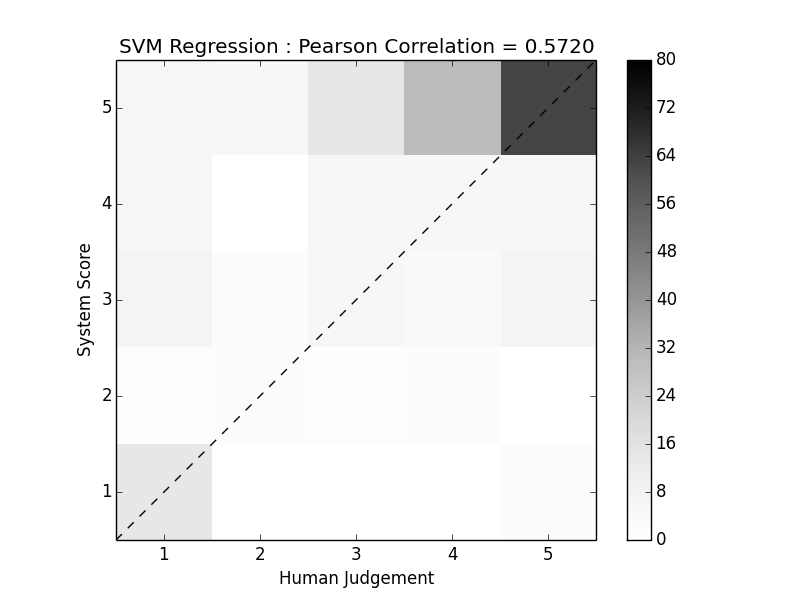}
\caption{Our Score \\ ($\rho$ = 0.5720)}
\label{fig:pearson_system_score} 
\end{subfigure}
\caption{Correlation between automatic scores (vertical axis) and 5-point human scores (horizontal axis) for ranking phrasal paraphrases. The darker squares along the diagonal line indicate a higher ranking.}
\label{fig:phrase_table_score_compare}
\end{figure*}


We used the GIZA++ word aligner in the Moses machine translation toolkit \cite{koehn2007moses} and extracted 245,686 phrasal paraphrases. Some examples are shown in Table \ref{tab:phrase_sample}. We additionally explored two supervised monolingual aligners: Jacana aligner \cite{yao-jacana-wordalign-acl2013} and Md Sultan's aligner \cite{sultan2014back}. We ranked the phrase pairs using four different scores: 

\begin{itemize}
    \item \textbf{Language Model Score} Let $w_{-2}w_{-1}pw_1w_2$ be the context of the phrase $p$. We considered a phrase $p'$ to be a good substitute for $p$ if $w_{-2}w_{-1}p'w_1w_2$ is a likely sequence according to a language model \cite{heafield2011kenlm} trained on Twitter data.
    \vspace{-.1in}
    \item \textbf{Translation Score} Moses provides translation probabilities $\varphi(p|p')$.
    \vspace{-.1in}
    \item \textbf{Glove Score} We used Glove \cite{pennington2014glove} pretrained 100-dimensional Twitter word vectors and cosine similarity.
    \vspace{-.1in}
    \item \textbf{Our Score} We trained a supervised SVM regression model using 500 phrase pairs with human ratings. We used the language model, translation, and glove scores as features, and additionally used the inverse phrase translation probability $\varphi(p'|p)$, lexical weighting $lex(p|p')$, and $lex(p'|p)$ from Moses.
\end{itemize}

Figure \ref{fig:phrase_table_score_compare} compares the different ranking methods against the human judgments on 200 phrase pairs randomly sampled from GIZA++.

\subsection{Paraphrase Quality Evaluation}

We compared the quality of paraphrases extracted by our method with the closest previous work (BUCC-2013) \cite{xu2013gathering}, in which a similar phrase table was created using Moses from monolingual parallel tweets that contain the same named entity and calendar date. We randomly sampled 500 phrase pairs from each phrase table and collected human judgements on a 5-point Likert scale, as described in Callison-Burch \cite{callison2008syntactic}. Table \ref{tab:phrase_table_compare} shows the evaluation results. We focused on the highest-quality paraphrases that rated as 5 (``all of the meaning of the original phrase is retained,
and nothing is added'') and their presence among all extracted paraphrases sorted by ranking scores. 

We were also interested in how these phrasal paraphrases compared with those in PPDB. We sampled an equal amount of 420 paraphrase pairs from our phrase tables and PPDB, and then checked what percentage out of the total 840 could be found in our phrase tables and PPDB, respectively. As shown in Table \ref{tab:phrase_coverage_compare}, there is little overlap between URL data and PPDB, only 1.3\% (51.3-50\%) plus 0.8\% (50.8-50\%). Our Twitter URL data complements well with the existing paraphrase resources, such as PPDB, which are primarily derived from well-edited texts.

\begin{table}[h]
\centering
\small
\resizebox{\columnwidth}{!}{%
\begin{tabular}{c|c|ccc}
\hline
\bf{Top Rankings}  & \bf{BUCC 2013}& \bf{GIZA++} &\bf{Jacana} &\bf{Sultan} \\ 
\hline
\bf{10\%} & 76.0 & 85.5 & 90.0 & 90.0\\ 
\bf{20\%} & 65.6 & 86.5 & 91.0 & 91.0\\ 
\bf{30\%} & 62.7 & 79.2 & 86.0 & 88.0\\
\bf{40\%} & 56.6 & 73.2 & 85.5 & 84.5\\
\bf{50\%} & 52.1 & 68.1 & 83.4 & 84.8\\
\bf{100\%} (all) & 36.3 & 49.8 & 75.8 & 77.2\\
\hline
\end{tabular}
}
\caption{\label{tab:phrase_table_compare} Percentage of high-quality phrasal paraphrases extracted from Twitter URL data (this work) by GIZA++, Jacana, Sultan aligners , comparing to the previous work (BUCC-2013).}
\end{table}

\begin{table}[h]
\centering
\resizebox{\columnwidth}{!}{%
\begin{tabular}{c|c|c||ccc}
\hline
 & \bf{PPDB} & \bf{URL} & \bf{GIZA++} & \bf{Jacana} & \bf{Sultan}  \\ \hline
\bf{Sample Size} & 50\% & 50\% & 16.7\% & 16.7\% & 16.7\% \\ 

\bf{Coverage} & 51.3\% & 50.8\% & 18.7\% & 32.1\% & 34.4\%   \\ \hline
\end{tabular}
}
\caption{Coverage comparison of phrasal paraphrases extracted from Twitter URL data (sampled 1:1:1 from GIZA++, Jacana and Sultan's aligner outputs) and the PPDB \cite{ganitkevitch2013ppdb}. } 
\label{tab:phrase_coverage_compare}
\end{table}

\section{Related Work}
\label{sec: related work}
\paragraph{Sentential Paraphrase Data} Researchers have found several data sources from which to collect sentential paraphrases: multiple news agencies reporting the same event (MSRP) \cite{dolan04,dolan2005automatically}, multiple translated versions of a foreign novel \cite{barzilay2003sentence,barzilay2003learning} or other texts \cite{cohn2008constructing}, multiple definitions of the same concept \cite{hashimoto2011extracting}, descriptions of the same video clip from multiple workers \cite{chen11} or rephrased sentences \cite{burrows2013paraphrase,toutanova-EtAl:2016:EMNLP2016}. However, all these data collection methods are incapable of obtaining sentential paraphrases on a large scale (i.e. limited number of news agencies or books with multiple translated versions), and/or lack meaningful negative examples. Both of these properties are crucial for developing machine learning models that identify paraphrases and measure semantic similarities. 


 \paragraph{Non-sentential Paraphrase Data}
There are other phrasal and syntactic paraphrase data, such as DIRT \cite{Lin:2001:DSI:502512.502559}, POLY \cite{grycner2016poly}, PATTY \cite{nakashole2012patty}, DEFIE \cite{bovi2015large}, and PPDB \cite{ganitkevitch2013ppdb,PavlickEtAl-2015:ACL:Semantics}. Most of these works focus on news or web data. Other earlier works on Twitter paraphrase extraction used unsupervised approaches \cite{xu2013gathering,blackparaphrasing} or small datasets \cite{zanzotto2011linguistic,antoniak2015leveraging}. 
\section{Conclusion and Future Work}
In this paper, we show how a simple method can effectively and continuously collect large-scale sentential paraphrases from Twitter. We rigorously evaluated our data with automatic identification classification models and various measurements. We will share our new dataset with the research community; this dataset includes 51,524 sentence pairs manually labeled and a monthly growth of 30,000 sentential paraphrases automatically labeled. Future work could include expanding into many different languages present in social media and developing language-independent automatic paraphrase identification models.

\section*{Acknowledgments}
We would like to thank Chris Callison-Burch, Weiwei Guo and Mike White for valuable discussions, as well as the anonymous reviewers for helpful feedback.

\bibliography{emnlp2017}
\bibliographystyle{emnlp_natbib}

\end{document}